\documentclass{article}
\usepackage{spconf,amsmath,graphicx}
\usepackage{amsmath}
\usepackage{amsfonts}
\usepackage{amssymb}

\usepackage[pagebackref,breaklinks,colorlinks]{hyperref}
\title{SAMF: Small-Area-Aware Multi-focus Image Fusion for Object Detection}
\name{Xilai Li, Xiaosong Li\sthanks{Corresponding author: Xiaosong Li (lixiaosong@buaa.edu.cn). This work was supported by the National Natural Science Foundation of China (Grant Nos. 62201149, 62271148, 62201151).}, Haishu Tan, Jinyang Li}
\address{School of Physics and Optoelectronic Engineering, Foshan University, Foshan 528225, China}
\begin{document}
%\ninept
%
\maketitle
\begin{abstract}
Existing multi-focus image fusion (MFIF) methods often fail to preserve the uncertain transition region and detect small focus areas within large defocused regions accurately. To address this issue, this study proposes a new small-area-aware MFIF algorithm for enhancing object detection capability. First, we enhance the pixel attributes within the small focus and boundary regions, which are subsequently combined with visual saliency detection to obtain the pre-fusion results used to discriminate the distribution of focused pixels. To accurately ensure pixel focus, we consider the source image as a combination of focused, defocused, and uncertain regions and propose a three-region segmentation strategy. Finally, we design an effective pixel selection rule to generate segmentation decision maps and obtain the final fusion results. Experiments demonstrated that the proposed method can accurately detect small and smooth focus areas while improving object detection performance, outperforming existing methods in both subjective and objective evaluations. The source code is available at \textcolor{red}{\href{https://github.com/ixilai/SAMF}{https://github.com/ixilai/SAMF}}.

\end{abstract}
\begin{keywords}
Multi-focus image fusion, small-area-aware, three-region segmentation, object detection
\end{keywords}
\section{Introduction}
\label{sec:intro}
The limited depth of field of optical lenses makes it difficult for optical imaging devices to satisfy the need for clear imaging of all objects in a scene, resulting in partially focused images. For some important practical applications, such as object detection, microscope imaging, and intelligent surveillance, all objects in an image must be in focus. This is achieved using a multi-focus image fusion (MFIF) technique \cite{R1,R3} that can fully extract focusing information from each source image. 

\begin{figure}[h]
  \centering
   \includegraphics[width=1.0\linewidth]{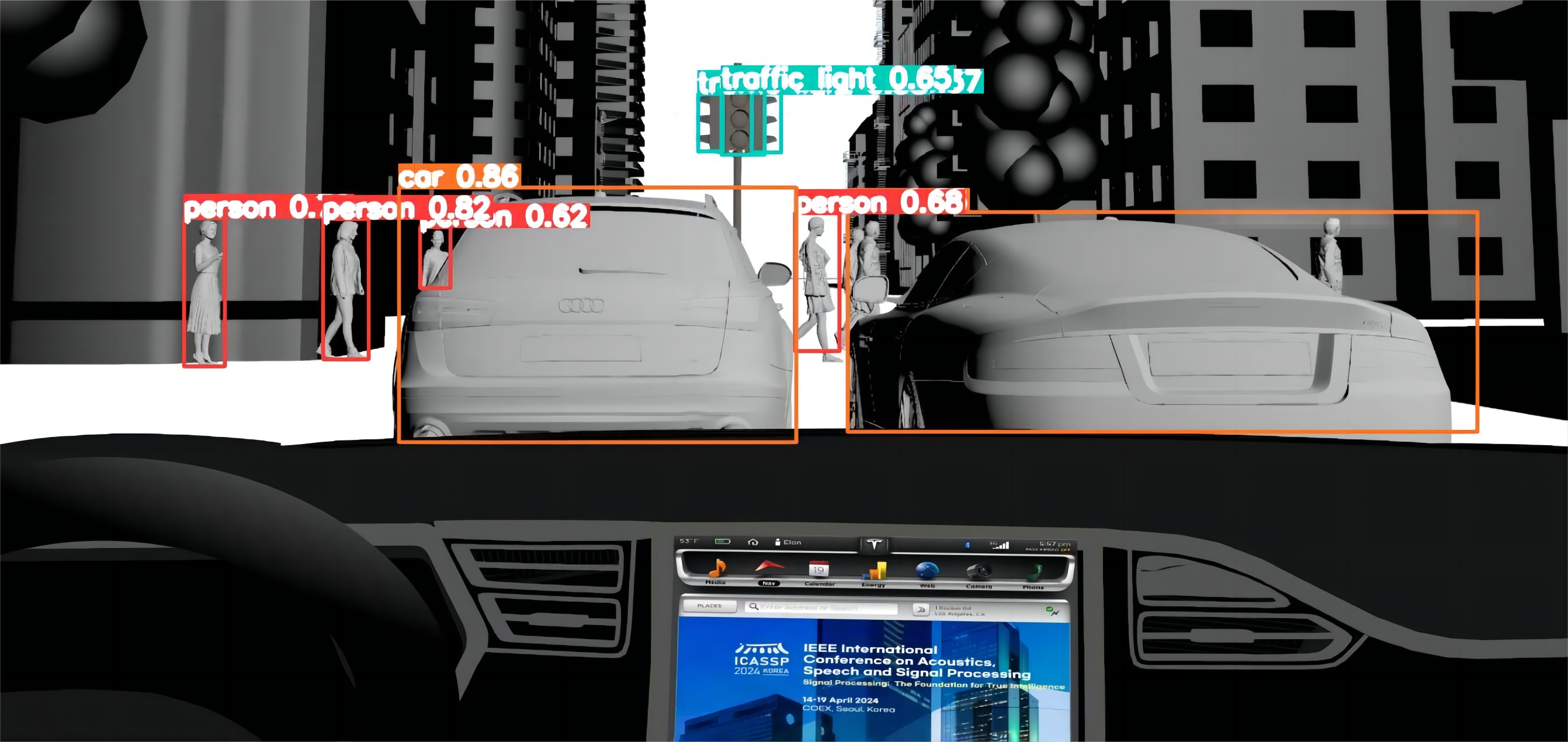}
   \caption{The application of MFIF in automatic driving.}
   \label{fig2}
\end{figure}
Existing image fusion algorithms can be divided into two main categories: deep learning (DL)-based methods \cite{R4,R5,R27} and traditional methods \cite{R7,R8,R9}. DL-based algorithms can be further categorized into decision map-based \cite{R4,R5,R10} and end-to-end-based \cite{R11,R12,R13} methods. The former utilize a decision map to determine the focusing attributes of pixels, allowing direct utilization of pixel information from the source image in the process of fusing multi-focus images, and the latter generate fusion images directly, without the need for postprocessing steps. The traditional methods can be further categorized into multi-scale transform (MST)-based \cite{R3,R14} and spatial domain-based \cite{R15,R16,R17} methods. MST-based methods mainly perform multi-scale decomposition of the source image, obtain a series of sub-bands representing different information from the source image, and finally design the fusion rules to integrate this information. Spatial domain-based algorithms construct a focusing decision map for fusion by detecting the saliency of pixels.

In summary, although existing methods can achieve desirable fusion performance, they may fail in recognizing small and complex focus regions, which are particularly important for object detection tasks. In real-world autonomous driving scenarios, it is difficult to capture the complete scene focus information within different depths of field in a single image. The goal of MFIF is to integrate the focus information from all depths of field into a unified image. However, owing to the complexity of road conditions, it is difficult for existing MFIF algorithms to extract object information accurately from compact background regions, such as pedestrians located between two vehicles; consequently, automobiles can fail to detect pedestrians, as shown in Fig. 1. To address this problem, we propose a small-area-aware MFIF algorithm (SAMF) for object detection. This study makes the following contributions: (1) We propose an innovative small-region-aware MFIF methods designed to address optical imaging constraints and improve object detection capability. (2) We propose a new three-region segmentation strategy that divides an image into focused, defocused, and uncertain regions. (3) We construct a real-world MFIF dataset (Road-MF) with all the images taken against vehicles on a road. The Road-MF dataset comprises 80 pairs of images, and fused images can be used for object detection tasks.

\begin{figure*}[h]
  \centering
  %\fbox{\rule{0pt}{2in} \rule{0.9\linewidth}{0pt}}
   \includegraphics[width=1.0\linewidth]{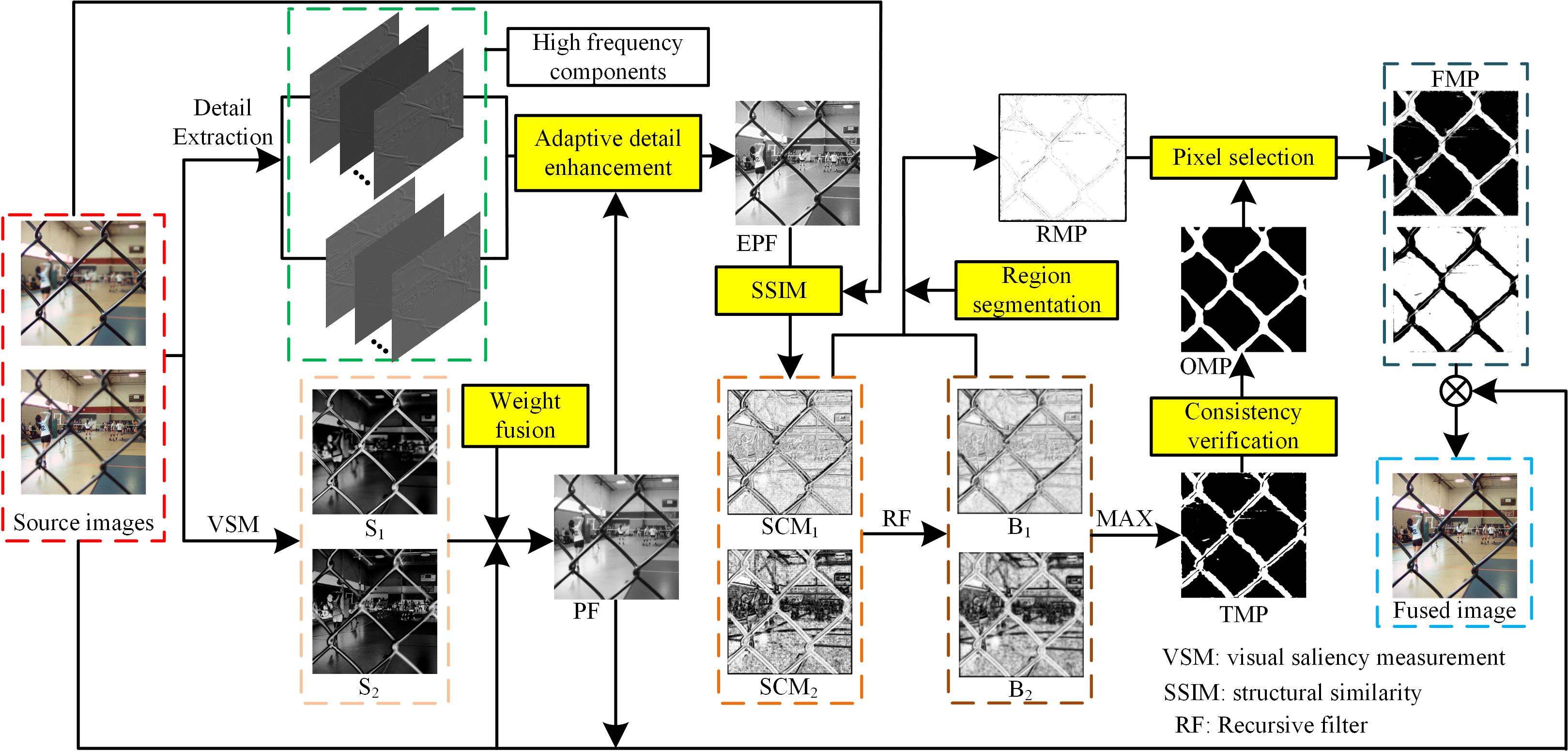}
   \caption{Flowchart of the proposed algorithm.}
   \label{fig2}
\end{figure*}

\section{Proposed method}
\label{sec:format}
We propose an SMAF method for object detection, and Fig. 2 shows the flowchart of this method.

\subsection{Enhanced Pre-fused Image Acquisition}
To effectively preserve information regarding details and texture, we develop a visual saliency measurement \cite{R18} (VSM)-based strategy to obtain the pre-fused result $PF$.
\begin{equation}
  PF(x, y) = W_F(x, y) \cdot I_1(x, y) + (1 - W_F(x, y)) \cdot I_2(x, y)
  \label{eq1}
\end{equation}
where $I_1$ and $I_2$ represent the multi-focus source images, the weight $W_F=0.5+(S_1-S_2)/2$ , $S_n$ represent the VSM value of the source image $I_n$, and $n \in \{1, 2, \ldots, N\}$. Subsequently, we utilize Gaussian filters to extract the detail information of the source image at different scales, and design an adaptive fusion rule based on log-energy. The specific algorithm is as follows:
\begin{equation}
  D_{m,n}(x,y) = I_n(x,y) - I_n(x,y) \ast G_{m,\sigma}
  \label{eq2}
\end{equation}
where $D_{m,n}(x,y)$ represents the high-frequency component of $I_n(x,y)$ at scale $m$, $*$ represents the convolution operator, $G_{(m,\sigma)}$ denotes the Gaussian function whose window size is $m \times m$, and $m \in \{1, 2, \ldots, M\}$. Afterwards, we superimpose all the high-frequency components $D_{m,n}(x,y)$  at different scales to obtain $\hat{D}_n(x, y)$, 
\begin{equation}
  \hat{D}_n(x,y) = \sum_{m=1}^M D_{m,n}(x,y)
  \label{eq3}
\end{equation}
Then, we compute log-energy $\hat{D}_n = \log(1 + \hat{D}_n^2)$ to analyze the degree of significant information differences between the different source images. Based on $E_n$, we can calculate the fused high-frequency component $FH$,
\begin{equation}
  FH =
\begin{cases}
  \sum_{n=1}^N HM_n \times \hat{D}_n,& |E_1 - E_2| > \lambda \\
  \sum_{n=1}^N W_h^n \times \hat{D}_n,& |E_1 - E_2| \leq \lambda
\end{cases}
  \label{eq4}
\end{equation}
where $HM_n$ represents the focused attribute decision map, which is generated by the pixel-wise maximum rule \cite{R9}, $W_h^n=E_n/(E_1+E_2)$ is the weighting coefficients of $\hat{D}_n$. For Eq. 3, we believe that when there are significant detail differences between different focusing regions, the pixel-wise maximum rule rule can effectively differentiate the focusing boundaries, and conversely, utilizing the weighted fusion rule will retain more useful information. Finally, we summed $FH$ with $PF$ to obtain the enhanced pre-fused result $EPF$.

\subsection{Acquisition of Regional Segmentation Decision Maps}
Many methods \cite{R3,R19} usually include only two outcomes for assessing pixel focus properties: focus or defocus. However, if these pixels are directly evaluated as focused or out of focus, this may result in some pixels within the small focus area not being fused. To overcome this issue, we propose a three-region segmentation strategy for accurate classification of pixels.

First, we extract information about the salient pixels within the focused region by calculating the structural similarity (SSIM) between the $EPF$ and the multi-focused source images. The SSIM scores for each pixel are summarized into a matrix to generate a score map $SCM_n$. Subsequently, we use recursive filter \cite{R20} (RF) to optimize the pixel classification in $SCM_n$ to obtain the two-region segmentation decision map $TMP$ by the following rule:
\begin{equation}
  TMP(x, y) =
\begin{cases}
1, & \text{if } \arg\max_n (B_1(x, y), \ldots, B_n(x, y)) \\
0, & \text{otherwise}
\end{cases}
  \label{eq5}
\end{equation}
where $B_n = RF(SCM_n)$, and $RF(\cdot)$ denotes the RF operation \cite{R20}. Finally, we use consistency verification technology \cite{R3} to process $TMP$ and obtain the optimized decision map $OMP$. This technology can scrutinize the coherence of the central pixel with the neighboring pixels within a fixed window to ascertain its placement in the in-focus or out-of-focus zone. The two-region segmentation strategy is able to recognize large focusing areas, however, this strategy may fail when pixel attribute judgment is performed in some small focusing areas and at focusing boundaries. Therefore, we propose a new three-region segmentation strategy.

First, we compute the difference map $DM = |SCM_1 - SCM_2|$ and then process it using RF to obtain difference blur map ($DBM$). Next, $B_n$ is used to obtain a blurred difference map ($BDM$), and $BDM=|B_1-B_2|$. Finally, $BDM(x,y)$ is compared with $DBM(x,y)$ to construct a decision map for three-region segmentation. Moreover, $BDM(x,y)$ can be infinitely close to $DBM(x,y)$ at the corresponding position only if the pixel is in focus or defocus. Otherwise, it is smaller than the $DBM(x,y)$ at the corresponding position. Using this feature, we construct the three-region segmentation decision map $RMP$ using the following pixel selection rules:
\begin{equation}
  RMP(x, y) =
\begin{cases}
  1, & \begin{aligned}
         &\text{if } BDM(x, y) > \beta \times DBM(x, y) \\
         &\text{and } B_1(x, y) > B_2(x, y)
       \end{aligned} \\
  2, & \begin{aligned}
         &\text{if } BDM(x, y) > \beta \times DBM(x, y) \\
         &\text{and } B_1(x, y) \leq B_2(x, y)
       \end{aligned} \\
  0.5, & \text{if } BDM(x, y) \leq \beta \times DBM(x, y)
\end{cases}
  \label{eq6}
\end{equation}
where $\beta=0.5$ is a balance parameter. For source image $I_1$, when $RMP(x,y)=1$, the pixel is considered a focusing pixel, when $RDM(x,y)=2$, it is regarded as a defocusing pixel, and it is regarded as an uncertain pixel when $RMP(x,y)=0.5$. For $I_2$, all cases are complementary except that the uncertain pixels are the same as for $I_1$. In addition, uncertain pixels are generally located at the boundary between the focusing and defocusing regions or in the surrounding region.

To combine the focus information captured by two-region and three-region segmentation strategies, we propose the following rule to obtain the final decision map ($FMP$):
\begin{equation}
FMP(x, y) =
\begin{cases}
  1, & \begin{aligned}
         &\text{if } OMP(x, y) = 1 \\
         & RMP(x, y) = 1
       \end{aligned} \\
  0, & \begin{aligned}
         &\text{if } OMP(x, y) = 0 \\
         & RMP(x, y) = 2
       \end{aligned} \\
  0.5, & \text{otherwise}
\end{cases}
  \label{eq7}
\end{equation}
Using  $FMP$ and $PF$, a focused, visually appealing, and fused image $F$ can be generated,
\begin{equation}
F(x, y) =
\begin{cases}
  I_1(x, y), & \text{if } FMP(x, y) = 1 \\
  I_2(x, y), & \text{if } FMP(x, y) = 0 \\
  PF(x, y), & \text{otherwise}
\end{cases}
  \label{eq8}
\end{equation}
The pixels that are judged to be in the focused or defocused region in $FMP$ are copied directly from the corresponding source image to $F$. Uncertain pixels are obtained from $PF$ to achieve smooth conversion from one source image to another.

\begin{figure}[h]
  \centering
  %\fbox{\rule{0pt}{2in} \rule{0.9\linewidth}{0pt}}
   \includegraphics[width=1.0\linewidth]{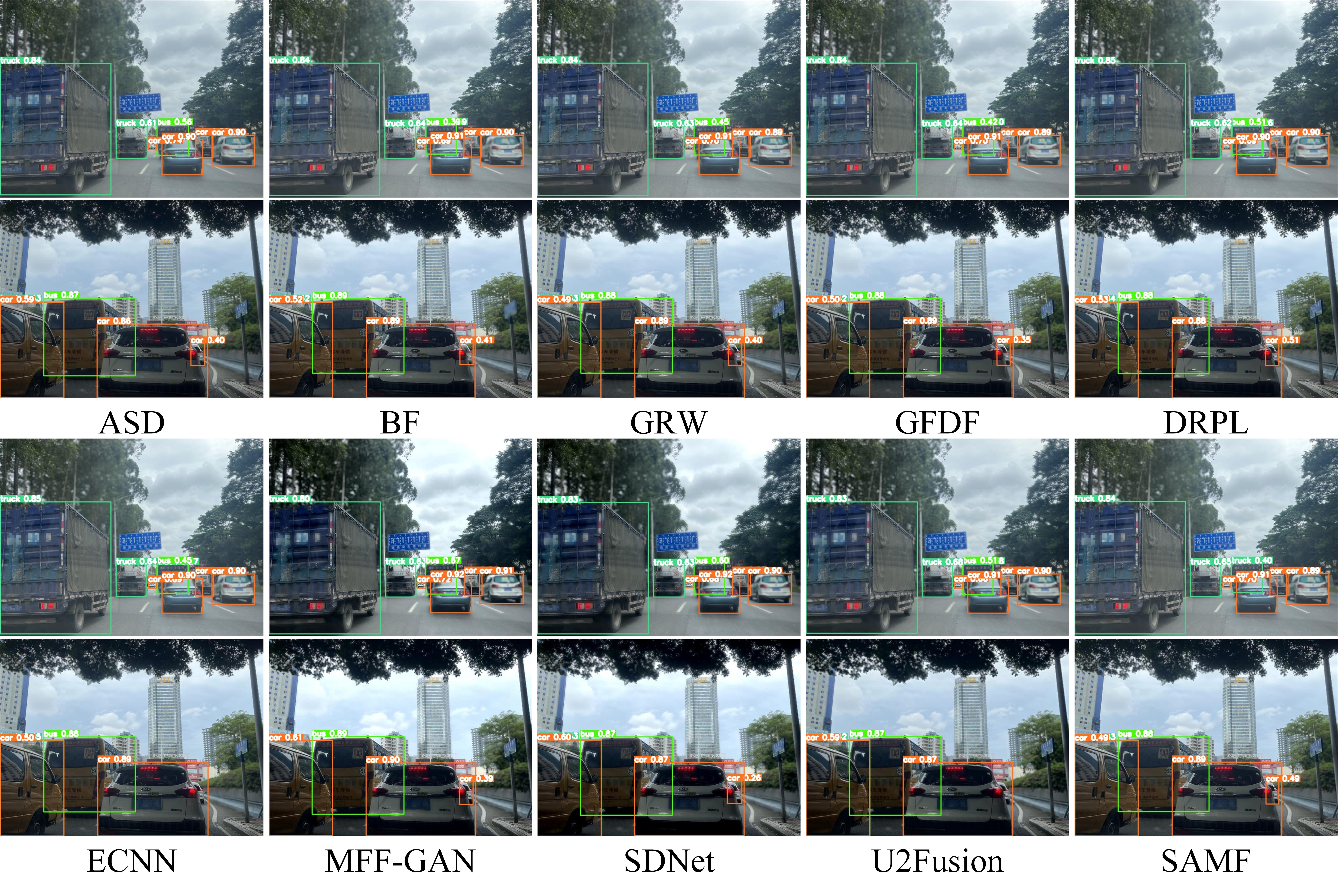}
   \caption{Fusion results of different methods on Road-MF dataset.}
   \label{fig3}
\end{figure}

\begin{figure*}[t]
  \centering
  %\fbox{\rule{0pt}{2in} \rule{0.9\linewidth}{0pt}}
   \includegraphics[width=1.0\linewidth]{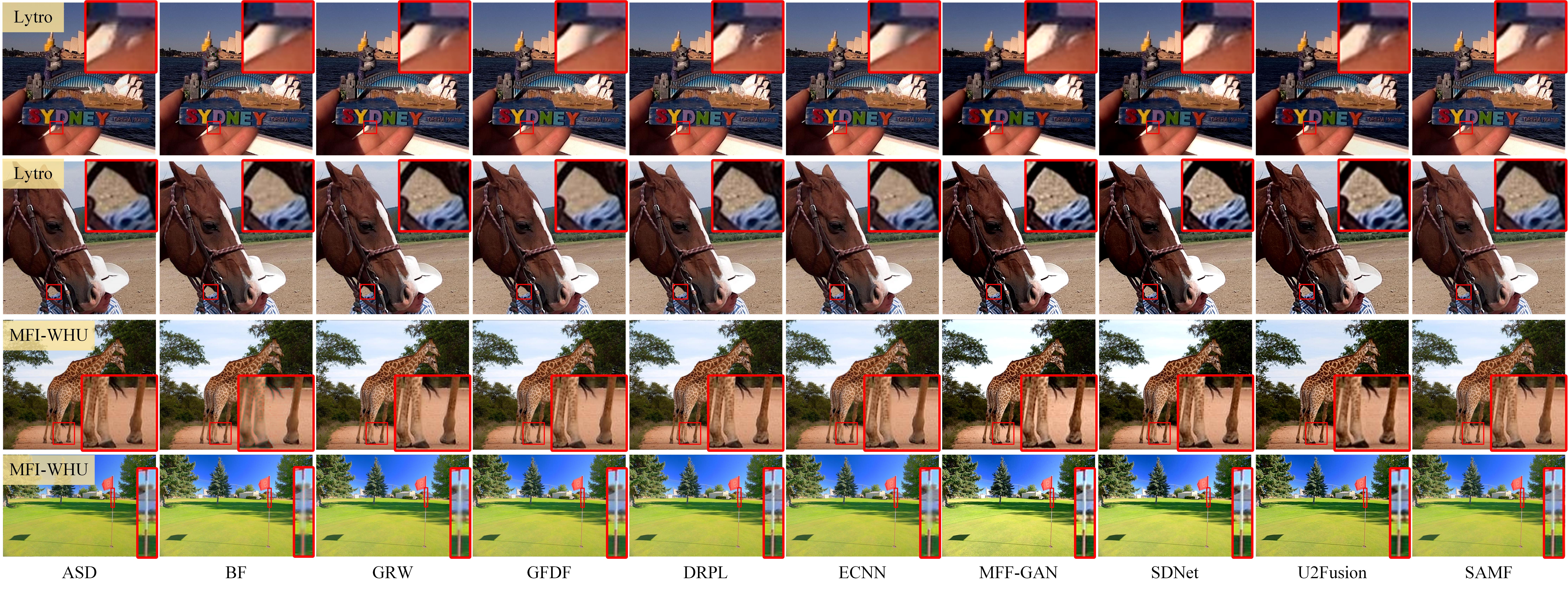}
   \caption{Fusion results of different methods on Lytro and MFI-WHU datasets.}
   \label{fig4}
\end{figure*}

\section{Experiment}
In this experiment, we qualitatively and quantitatively compare the SAMF with nine state-of-the-art comparative methods in three datasets, Lytro \cite{R21}, MFI-WHU \cite{R22} and proposed Road-MF datasets. Nine comparison methods were used: ASD \cite{R23}, BF \cite{R16}, GRW \cite{R15}, GFDF \cite{R9}, DRPL \cite{R5}, ECNN \cite{R4}, MFF-GAN \cite{R22}, SDNet \cite{R24}, and U2Fusion \cite{R11}. In addition, we selected four objective evaluation metrics, including: $Q_{MI}$, $Q_M$, $Q_S$, and $Q_{CV}$ \cite{R25}. Except for with $Q_{CV}$, higher scores represent better fusion performance.

\begin{table*}
\centering
\small % 设置表格内字体为更小一号
\caption{Quantitative comparison results of all methods on three datasets. \textbf{Bold} is the best.}
\label{tab1}
\begin{tabular}{|l|c|c|c|c|c|c|c|c|c|c|c|c|}
\hline
\textbf{}  & \multicolumn{4}{c|}{\textbf{Lytro dataset}} & \multicolumn{4}{c|}{\textbf{MFI-WHU dataset}} & \multicolumn{4}{c|}{\textbf{Road-MF dataset}} \\
\hline
 & $Q_{MI}$ & $Q_M$ & $Q_S$ & $Q_{CV}$ & $Q_{MI}$ & $Q_M$ & $Q_S$ & $Q_{CV}$ & $Q_{MI}$ & $Q_M$ & $Q_S$ & $Q_{CV}$ \\
\hline
ASD  & 0.9703 & 0.8448 & 0.9464 & 24.9008 & 0.9739 & 0.8351 & 0.9468 & 48.7762 & 1.0667 & 0.7048 & 0.9510 & 13.6782 \\
BF  & 1.1668 & 2.4381 & 0.9452 & 20.2330 & 1.1937 & 2.3735 & 0.9453 & 25.0691 & 1.1135 & 0.8582 & 0.9506 & 5.5447 \\
GRW  & 1.1601 & 2.4286 & \textbf{0.9466} & 16.5851 & 1.1797 & 2.3636 & \textbf{0.9495} & 30.3327 & 1.1083 & 0.8781 & 0.9521 & 5.2630 \\
GFDF  & 1.1643 & 2.5020 & 0.9463 & 15.8883 & 1.1898 & 2.4345 & \textbf{0.9495} & 29.3793 & 1.1095 & 0.8841 & 0.9520 & 4.9977 \\
DRPL  & 1.0915 & 1.7030 & 0.9445 & 16.7754 & 1.1041 & 1.6716 & 0.9483 & 29.8766 & \textbf{1.2800} & \textbf{1.7004} & 0.9606 & 3.7082 \\
ECNN  & 1.1318 & 2.2030 & 0.9423 & 16.5572 & 1.1920 & 2.4462 & 0.9476 & 31.2840 & 1.1097 & 0.8888 & 0.9515 & 5.1107 \\
MFF-GAN  & 0.8343 & 0.5834 & 0.8752 & 67.8460 & 0.7832 & 0.4348 & 0.8830 & 109.0935 & 0.9779 & 0.4563 & 0.9236 & 22.2484 \\
SDNet  & 0.8343 & 0.5834 & 0.8752 & 67.8460 & 0.8738 & 0.5297 & 0.9006 & 55.2413 & 1.0893 & 0.6705 & 0.9426 & 7.2748 \\
U2Fusion  & 0.7989 & 0.4699 & 0.8943 & 49.4947 & 0.7302 & 0.4053 & 0.8973 & 58.1453 & 0.9348 & 0.3897 & 0.9459 & 28.0869 \\
SAMF & \textbf{1.1781} & \textbf{2.5991} & 0.9459 & \textbf{15.7835} & \textbf{1.2155} & \textbf{2.5165} & 0.9487 & \textbf{21.7840} & 1.1095 & 0.8820 & \textbf{0.9619} & \textbf{3.0818} \\
\hline
\end{tabular}
\end{table*}

\noindent\textbf{\textit{Qualitative comparison of object detection performance}}
Fig. 3 shows the fusion results of the proposed algorithms and the comparison methods on the Road-MF dataset. We utilized YoloV5 \cite{R26}, an object detection algorithm,  to test all the fu- sion results and evaluate their fusion performance. As shown in the figure, all seven algorithms (ASD, BF, GRW, GFDF, ECNN, SDNet and U2Fusion) had low detection accuracy for some small area objects. In addition, all comparison methods except the proposed algorithm suffered from the object misclassification problem.

\noindent\textbf{\textit{Qualitative comparison of fusion performance}}
Fig. 4 shows the qualitative comparison results of the proposed algorithm and the nine compared methods on the Lytro and MFI-WHU datasets. As can be seen from the red zoomed-in area, BF, GRW, GFDF, and ECNN did not preserve the pixel information from the small focused area. ASD, MFF-GAN, SDNet, and U2Fusion lost contrast and sharpness information from the focused area, and DRPL was unnatural in processing the pixels at the focusing boundaries and showed residual artifacts. Since SAMF precisely divides the source image into three regions for fusion, it could effectively detect the small-area blocks in the focused region and provide high quality fusion results. 

\noindent\textbf{\textit{Quantitative Comparison}}
Table 1 shows the quantitative comparison results between the proposed algorithm and nine comparison methods on three datasets. The results show that the proposed algorithm achieved the best scores in terms of most of the metrics, which demonstrates the effectiveness of our proposed three-region segmentation strategy, which can effectively retain the focusing information from different source images and has better contrast and clarity compared with the nine comparison methods.
\section{Conclusion}
In this study, we proposed SAMF for object detection. Our method excels at detecting small focus regions and precisely determining the pixel focus characteristics in the vicinity of boundaries. The proposed model treats the source image as a combination of focused, out-of-focus, and uncertain regions. Experimental results demonstrated that the proposed method outperformed some state-of-the-art techniques. In future work, we will focus on solving the small-area MFIF problem when the source images have different resolutions and are unaligned so that our research van effectively be extended to intelligent recognition tasks.

% References should be produced using the bibtex program from suitable
% BiBTeX files (here: strings, refs, manuals). The IEEEbib.bst bibliography
% style file from IEEE produces unsorted bibliography list.
% -------------------------------------------------------------------------
\bibliographystyle{IEEEbib}
\small % 设置参考文献字体为小一号
\bibliography{strings,refs}

\end{document}